\documentclass[letterpaper, 10pt, conference]{ieeeconf}  	

\IEEEoverridecommandlockouts

\usepackage{times}
\usepackage{tabularx}
\usepackage{subcaption}
\usepackage{multirow,multicol, array}
\usepackage[font={small}]{caption}   
\usepackage{graphicx,dblfloatfix}
\usepackage{wrapfig}
\usepackage{siunitx}
\usepackage{soul}
\usepackage{marginnote}

\let\labelindent\relax
\usepackage{enumitem}

\usepackage{amsmath,amsthm,amssymb,amsfonts}
\usepackage[titlenumbered,ruled]{algorithm2e}
\usepackage{textcomp}
\usepackage{acronym}
\usepackage{balance}
\usepackage{mdwmath}
\usepackage{bm}
\usepackage{mdwtab}
\usepackage{array}
\usepackage{eqparbox}
\usepackage{cite}
\usepackage{verbatim}
\usepackage{amssymb}

\usepackage{color}
\usepackage{xcolor}
\usepackage{psfrag}
\usepackage{epsfig}
\usepackage{url}

\usepackage{epstopdf}
\usepackage{booktabs}
\usepackage{blindtext}

\usepackage{physics}

\usepackage[colorinlistoftodos,prependcaption,textsize=tiny]{todonotes}

\renewcommand{\emph}{\textit}

\newtheorem*{lemma*}{Lemma}

\newtheorem*{problem*}{Problem}

\makeatletter
\newcommand\fs@spaceruled{\def\@fs@cfont{\bfseries}\let\@fs@capt\floatc@ruled
	\def\@fs@pre{\vspace{5\baselineskip}\hrule height.8pt depth0pt \kern2pt}%
	\def\@fs@post{\kern2pt\hrule\relax}%
	\def\@fs@mid{\kern2pt\hrule\kern2pt}%
	\let\@fs@iftopcapt\iftrue}
\def\maketag@@@#1{\hbox{\m@th\normalfont\normalsize#1}}
\makeatother

\IEEEoverridecommandlockouts
\graphicspath{{images/}}

\definecolor{darkgreen}{rgb}{0.1, 0.6, 0.1}

\newcommand{\mycomment}[1]{}

\title{Design and Central Pattern Generator Control of a New Transformable Wheel-Legged Robot} 
    
\author{Tyler Bishop, Keran Ye, and Konstantinos Karydis
\thanks{The authors are with the Dept. of Electrical and Computer Engineering, University of California, Riverside.
Email: {\{tbish006, kye007, karydis\}@ucr.edu}.
}
\thanks{
We gratefully acknowledge the support of NSF \# CMMI-2046270, ARL \# W911NF-18-1-0266, and The University of California \# UC-MRPI M21PR3417. Any opinions, findings, and conclusions or recommendations expressed in this material are those of the authors and do not necessarily reflect the views of the funding agencies.}
}

\begin{document}

\maketitle
\thispagestyle{empty}
\pagestyle{empty}


\begin{abstract}
This paper introduces a new wheel-legged robot and develops motion controllers based on central pattern generators (CPGs) for the robot to navigate over a range of terrains. A transformable leg-wheel design is considered and characterized in terms of key locomotion characteristics as a function of the design. Kinematic analysis is conducted based on a generalized four-bar mechanism driven by a coaxial hub arrangement. The analysis is used to inform the design of a central pattern generator to control the robot by mapping oscillator states to wheel-leg trajectories and implementing differential steering within the oscillator network. Three oscillator models are used as the basis of the CPGs, and their performance is compared over a range of inputs. The CPG-based controller is used to drive the developed robot prototype on level ground and over obstacles. Additional simulated tests are performed for uneven terrain negotiation and obstacle climbing. Results demonstrate the effectiveness of CPG control in transformable wheel-legged robots.
\end{abstract}


\vspace{-1pt}
\section{Introduction}
\label{intro}


Locomotion over unstructured terrain is a challenge to address when developing mobile robots. While legged and wheeled robots are most common, there have been several efforts over the years to develop and deploy in applications wheel-legged hybrid robots (e.g.,~\cite{jeans2009impass,zarrouk2013star,karydis2014planning,she2015transformable,stager2015passively,eich2009proprioceptive,lambrecht_small_2005}).
These include designs and methods which attempt to combine the versatility of legged locomotion to overcome obstacles with the efficiency and ease of control afforded by wheeled locomotion.
One such method, which is most relevant to this paper, considers wheels that can in some way transform their shape to be more able to navigate over challenging terrain \cite{tadakuma2010mechanical,lee2014fabrication,mertyuz2020fuhar,woolley2021cylindabot,cao2022omniwheg,yun2017development}.

The ability to operate in different modes~\cite{ye2023novel,ye2021modeling} is especially well suited to semi-natural environments such as agricultural areas \cite{chen2013quattroped,chen2017turboquad}. Managed semi-natural land often consists of primarily flat ground that can be easily traversed by wheeled vehicles. However, occasional large obstacles such as fallen tree branches and furrowed or muddy terrain can pose a problem for robots with smaller wheels. 
Large wheels used on heavier robotic systems could damage fragile crops and irrigation equipment. Some agricultural inspection robots, such as the commercial unit SentiV~\cite{SentiVMeropy}, use rimless wheel design to reduce this potential for damage. For fields with large navigable empty spaces, a transformable wheel may be more efficient.

A common leg-wheel mechanism in the literature uses a motor and four bar mechanism assembly inside the wheel to provide the actuation force for transformation \cite{cao2022omniwheg,mertyuz2020fuhar}. Other methods involve passively transforming wheels \cite{kim2014wheel,zheng2019wheeler}. In contrast, in this work the mechanism is driven by a coaxial shaft arrangement. This allows the motors to be located outside of the wheel while still allowing full control of the state of the wheel and its internal mechanism.

In addition, navigation in semi-natural environments requires a controller that can direct the robot's movement over a variety of obstacles and terrains while exhibiting consistent behavior and smooth transitions between modes. Analyses of transforming leg-wheels have often focused on traversing vertical step-like obstacles~\cite{cao2022omniwheg,eich2009proprioceptive,bai_wheel-legged_2021}. The torque requirements of motors and the precise body trajectories of the robot during obstacle climbing have been well quantified~\cite{mertyuz2020fuhar,cao2022omniwheg,kim2020step}. An approach to designing a controller for natural environments is proposed in \cite{chen2017turboquad} and \cite{eich2009proprioceptive} where a central pattern generator (CPG) allows for smooth switching between wheeled and legged locomotion as well as the option of driving the leg-wheels with specific gaits. The use of CPGs has proved very effective for control of many locomotion strategies and can be robust for a variety of situations \cite{pinto2006central,righetti2006design,ijspeert2008central,crespi2008controlling,liu2009cpg,garcia2015central,righetti2008pattern,ijspeert2005simulation,ijspeert_swimming_2007}.
\begin{figure}[!t]
\vspace{6pt}
\centering
\includegraphics[width=0.475\textwidth]{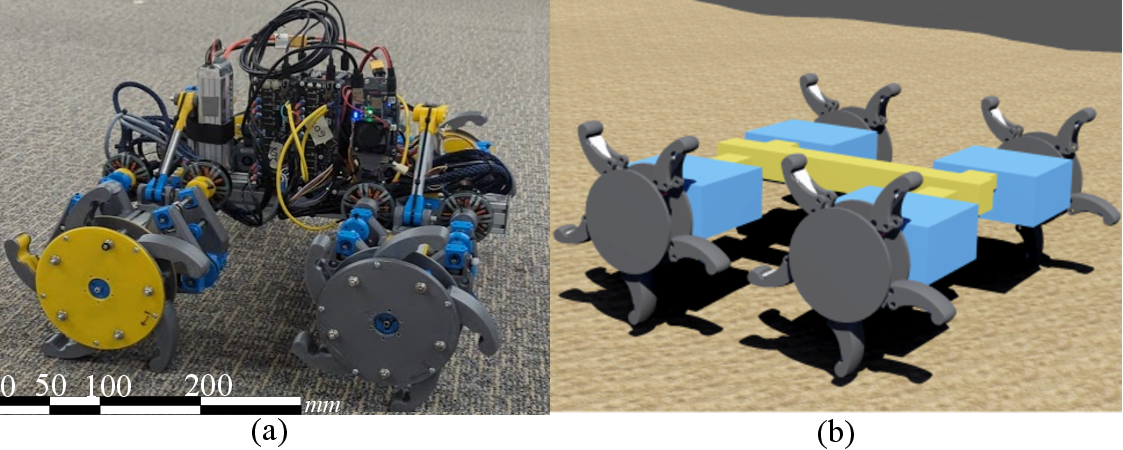}
\caption{ (a) The robot prototype has four transforming leg-wheels driven by eight BLDC motors and arranged to use differential steering. (b) The robot is modeled in Webots to perform simulated experiments on various obstacles and terrain}
\label{f:robot}
\vspace{-15pt}
\end{figure}

A challenge presented when using CPG control for wheeled robots is its integration with the steering mechanism of the robot. In agricultural robotics, differential steering is common for four wheeled robots \cite{gaoReviewWheeledMobile2018,clearpathHusky}.
This is mechanically simpler to implement than Ackerman or articulated steering but requires modification of the CPG network to drive the wheels at precise velocities.
An additional challenge is to provide smooth movement when the robot is driven in legged mode, which requires the legs to be periodically retracted to maintain the same effective radius. This is important when performing inspection tasks where sensor stability is required, as well as for reducing the dynamic loads on the robot and drive motors. CPGs offer a solution to this problem, by producing periodic outputs that are closely coupled with their own internal phase.

The use of transforming leg-wheels for mobile robots operating in agricultural environments requires robust control over multiple modes of locomotion and unpredictable terrain. Central pattern generators have been used successfully in similar contexts. To this end, this paper seeks to expand their application to agricultural robots with an emphasis on uneven terrain and the use of differential steering.
Key contributions include 1) the design and the geometric and kinematic analysis of a new leg-wheel mechanism with a focus on optimizing for obstacles present in an agricultural context, and 2) the development and testing of different oscillator models in the design of a central pattern generator to control the robot. Our developed robot prototype is tested using the CPGs in a variety of real and simulated environments.

\section{Mechanical Design and Kinematics}
\label{Kinematic}

\subsection{Prototype design and construction}



\begin{figure}[!t]
\vspace{6pt}
\centering
\includegraphics[width=0.5\textwidth]{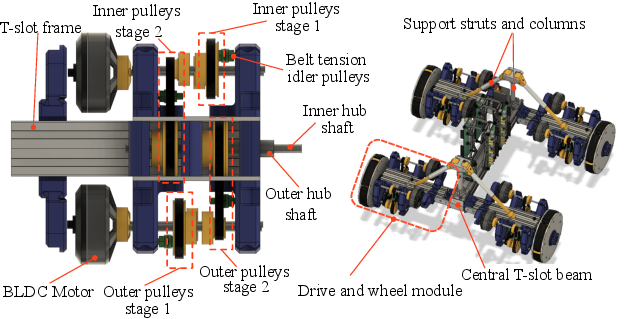}
\caption{CAD design of the drive section and full robot assembly. The drive uses two two-stage belt and pulley systems for power transmission. Different sized pulleys can be 3D-printed and swapped in without re-fabricating the structural elements. The inner wheel hub is driven through a 3D-printed spline; the outer wheel hub is attached rigidly to the outer shaft.}
\vspace{-10pt}
\label{f:cad}
\end{figure}

The robot is designed with four drive and wheel modules attached to a central beam and stiffened with supporting struts from two central columns (Fig.~\ref{f:cad}). The robot frame uses T-slot aluminum extrusions to allow the wheel modules and other components to be relocated along the central axis of the robot.
To avoid the need for a rotating electrical connection which is commonly employed in related works~\cite{cao2022omniwheg,mertyuz2020fuhar}, the drive mechanism in this work was designed to use two coaxial shafts, each driven by an external motor.

The configuration of the wheel-leg mechanism was determined based on this work's motivating application, i.e. navigation in drip-irrigated agricultural fields. To this end, we performed a cursory examination of real terrain morphology at the University of California Riverside (UCR) Agricultural Experimental Station (AES) to identify typical obstacles present in this context. These are summarized in Table~\ref{t:obs}.

A geometric analysis of leg-wheel shapes is needed in order to utilize the obstacle information reported in Table~\ref{t:obs} for robot design. A general, simplified wheel shape was considered where the circumference of the wheel is divided into $n$ equal arcs (Fig.~\ref{f:wheel}) with arc angle $\alpha=\frac{2\pi}{n}$. A key design consideration is the number of legs (arcs) within each wheel since this determines the ratios between wheel radius and important locomotion parameters such as step height. The leg length, $L_{arc}$, and step length, $L_{step}$, are calculated from radius $r$ as
\begin{equation}
\label{eq:geo1}
	L_{arc} = \sqrt{2 r^2 \cdot (1-\cos(\alpha)},\hspace{5pt} L_{step} = 2(r+L_{arc})\sin(\frac{\alpha}{2})\;.
\end{equation}
The height of the wheel center when legs are fully extended can be found from leg length, radius, and angle $\alpha$ as
\begin{equation}
\label{eq:geo2}
	h_{min} = (r+L_{arc}) \cdot cos(\frac{\alpha}{2}), \hspace{5pt} h_{max} = r+L_{arc}\;.
\end{equation}

Computed values for these variables for different numbers of legs are given in Table \ref{t:geo}. The minimum center height is maximized for a wheel with $n=5$ arcs. This variable is considered the most important for our application to allow the robot to step over irrigation lines without damaging them.

\begin{table}[!tb]
\vspace{6pt}
	\caption{Obstacle Dimensions in Agricultural Fields}
	\vspace{-9pt}
	\label{t:obs}
	\begin{center}
	\renewcommand{\arraystretch}{1.5}
	\begin{tabular}{p{2.5cm}|p{1.5cm}>{\centering}p{1.4cm}>{\centering}p{1.4cm}}
	\toprule
	Obstacle & Dimension & Min (mm) & Max (mm)\\
	\midrule
	Irrigation pipe & diameter & 50 & 110\\
	Ditch/Berm & width & 150 & 300\\
	Ditch/Berm & depth/height & 50 & 170\\
	Low tree branches & height & 450 & N/A\\
	\bottomrule
   	 
	\end{tabular}
	\end{center}
	\vspace{-18pt}
\end{table}

\begin{table}[!tb]
\vspace{9pt}
	\caption{Key Leg-Wheel Variables for N Legs}
	\vspace{-9pt}
	\label{t:geo}
	\begin{center}
	\renewcommand{\arraystretch}{1.5}
	\begin{tabular}{ p{2.6cm} | p{0.5cm} p{0.5cm} p{0.5cm} p{0.5cm} p{0.5cm} p{0.5cm} }
	\toprule
	Variable $\backslash$ \# Arcs & 3 & 4 & 5 & 6 & 7 & 8 \\
	\midrule
	$L_{step}$ & 4.73 & 3.41 & 2.56 & 2.00 & 1.62 & 1.35 \\
	$h_{min}$ & 1.37 & 1.71 & 1.76 & 1.74 & 1.68 & 1.63 \\
	$h_{max} - h_{min}$ & 1.37 & 0.71 & 0.42 & 0.27 & 0.18 & 0.13 \\
	\bottomrule
   	 
	\end{tabular}

	\end{center}
	\footnotesize{Values are unit-less, relative to a radius of $r=1$}
	\vspace{-15pt}
\end{table}

\subsection{Kinematic and Quasi-static Analysis} \label{kinematic}

A simplified model of the wheel leg mechanism is used to derive the inverse kinematics for a wheel using a generic four-bar mechanism.
Inverse kinematics is solved by twice applying the general equations of a two-link robotic arm given by \eqref{eq:IK}, which can be derived from Ch 6 intro of \cite{lynchModernRoboticsMechanics2017}
\begin{align}
\theta_2 &= -\cos^{-1}(\frac{x^2 + y^2 - L_1^2 - L_2^2}{2\cdot L_1 \cdot L_2}\;), \nonumber \\
\theta_1 &= \tan^{-1}(\frac{y}{x}) - \tan^{-1}(\frac{L_2 \cdot \sin(\theta_2)}{L_1 + L_2\cdot \cos(\theta_2)})\;.
\label{eq:IK}
\end{align}

\begin{figure*}
\vspace{6pt}
\centering
\includegraphics[width=1.0\textwidth]{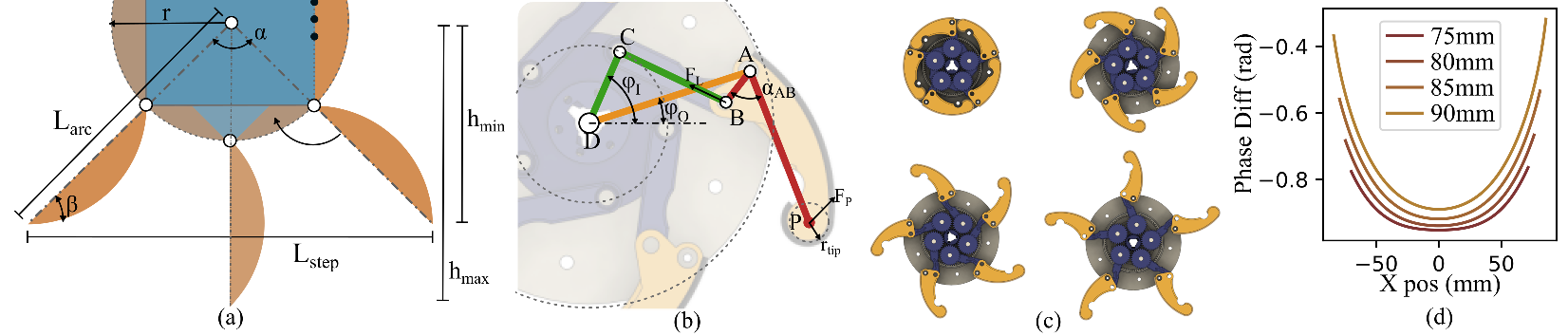}
\vspace{-10pt}
\caption{(a) Wheel geometry relative to the number of legs (arcs). (Here, a wheel with $n=4$ legs is shown.) (b) The links of the four bar mechanism are made up of the inner hub $\vec{DC}$, outer hub $\vec{DA}$, wheel arc  $\vec{AB}$, and a bar link $\vec{CB}$. (c) Planetary gear mechanism added in later iterations which uses the same four bar mechanism but with the inner hub replaced by the planet gear carrier. (d) U-shaped trajectory of the phase offset between wheel hubs relative to a tip position with a constant Y component.}
\label{f:wheel}
\vspace{-20pt}
\end{figure*}

The first virtual arm is made from the wheel arc and outer hub (Fig.~\ref{f:wheel}(b)), thus $L_2 = \overline{AP}$, $L_1 = \overline{DA}$. Equation \eqref{eq:IK} is solved, with $x,y$ being the arc tip position relative to the wheel center, to give the position of joint $A$ and the outer hub phase angle $\phi_O = \theta_1$. These are then used to find the position of joint B,

 \begin{equation}\label{eq:IK3}
	\begin{bmatrix}
	B_x \\ B_y
	\end{bmatrix} =
	\begin{bmatrix}
	A_x + \overline{AB} \cdot \cos(\phi_1 + \alpha_{AB})\\
	A_y + \overline{AB} \cdot \sin(\phi_1 + \alpha_{AB})
	\end{bmatrix}\;.
\end{equation}
The second virtual arm is made by the inner hub and link, $L_1=\overline{DC}$, $L_2=\overline{CB}$. Equation \eqref{eq:IK} is solved again using the $x,y$ position of point B to find inner hub phase $\phi_{I} = \theta_1$.

Further analysis was done to characterize the mapping between the position of the leg tip and the phases of the inner and outer hub. For the case of the leg tip moving across the ground with constant velocity and height, the relative phase follows a U-shaped trajectory (Fig. \ref{f:wheel}). The precise shape depends on the wheel design and desired height. The shape of this curve informed the choice of a CPG-based controller, given its periodicity with every step (rotation of $\frac{\pi}{N}$ rad).

Output torques on the driven hubs of the wheel were computed assuming quasi-static conditions. The wheel components' inertia and weight are much smaller than the robot as a whole. Thus, the loading was analyzed assuming a single force vector $\vec{F_P}$ acting on the tip of the arc $P$. Let $\hat{l}$ denote the unit vector along the link member $\vec{BC}$ and $F_L$ be the tension in the member. Static analysis of the moments acting on the wheel arc about point $A$ yields
 	$\vec{AB} \times \left( F_L \cdot \hat{l} \right) = \vec{AP} \times \vec{F_P}$.

This can be used to directly calculate $F_L$. This force is applied to the inner hub and can be used to find the torque acting about the center pivot as  
	$\tau_I = || \vec{DC} \times \left( F_L \cdot \hat{l} \right)||$. 
%
The torque on the outer hub is produced entirely by the reaction forces acting on the wheel arc at point A; thus, 
	$\tau_O = || \left( F_L \cdot \hat{l} + \vec{F_P}  \right) \times \vec{DA}||$.
\vspace{-1pt}


Preliminary testing suggested the need to integrate a planetary gear system. The goal is to increase the phase difference needed to extend the wheel completely, thereby reducing the effects of flexing in the drive section. The inverse kinematics are applied exactly as before with the planetary carrier acting as the inner hub $\vec{CD}$. However, this also changes the torque applied to the inner and outer hubs as
	$\tau_{Op} = \tau_O + \tau_I \cdot \frac{N_R}{N_R+N_S}, \hspace{5pt}
	\tau_{Ip} = \tau_I \cdot \frac{N_S}{N_R+N_S}$. 
This results in a much higher loading on the outer hub motor, but is still within an acceptable range for the chosen hardware.
The design parameters for the wheel mechanism used on our prototype robot are listed in Table~\ref{t:wheel_par}.

\section{Controller Design}
\label{controller}
\subsection{Controller Architecture}
As discussed in Section \ref{kinematic} leg-wheel motion with a constant height and velocity requires a constant phase trajectory with a U-shaped periodic component added to the inner hub phase. To provide this signal, a CPG constructed from oscillators with a well defined limit cycle phase is considered. These are commonly used in CPG control of quadruped robots, with a focus on inter-oscillator phase relationships. Three commonly used models are the Kuramoto, Hopf, and Van der Pol oscillators~\cite{mombaur2017control,ijspeert2008central}. 

\begin{table}[!tb]
\vspace{6pt}
	\caption{Table of Wheel Leg Parameters}
	\vspace{-9pt}
	\label{t:wheel_par}
	\begin{center}
	\renewcommand{\arraystretch}{1.5}
	\begin{tabular}{p{1.5cm}>{\centering}p{1.5cm}>{\centering}p{0.6cm}p{3.2cm}}
	\toprule
	Parameter & Value & Unit & Definition \\
	\midrule
	$\overline{AB}$ & 15 & mm & Wheel arc pivot radius\\
	$\overline{DA}$, $\overline{DC}$ & 65.0, 28.0 & mm & Outer, Inner hub radius\\
	$\overline{CB}$, $\overline{AP}$ & 45.3, 62.3 & mm & Link Member, Arc length\\
	$r_{p}$ & 8 & mm & Arc tip radius\\
	$\alpha_{AB}$ & 59.4 & deg & Angle between $\overline{AB}$ and arc\\
	$N_S$,$N_R$,$N_P$  & 24, 29, 82 & - & Sun, Planet, Ring gear teeth\\
	\bottomrule
   	 
	\end{tabular}
	\end{center}
	\vspace{-25pt}
\end{table}

\begin{figure}[!tb]
\vspace{6pt}
\centering
\includegraphics[width=0.475\textwidth]{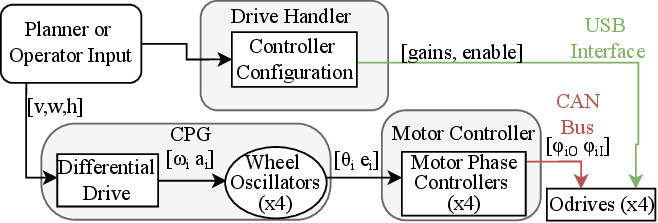}
\caption{The controller runs on a BeagleBone Blue using ROS with separate nodes (grey boxes). Motor control is provided by four Odrive BLDC boards which employ cascaded PID controllers when running in position control mode \cite{OdriveController}. Position commands and estimates are sent between the Beagle Bone and Odrive controllers over CAN Bus. Other control parameters are sent over USB.}
\label{f:controller}
\vspace{-18pt}
\end{figure}

Each wheel is controlled by a single oscillator, with the phase of the oscillator taken as the control variable for rotational position $\theta_i$. A secondary output $e_i$ is constructed to command the phase offset between the wheel hubs. Thus, and with reference to Fig.~\ref{f:wheel}, $\phi_{O_i} = \theta_i$ and $\phi_{I_i} = \theta_i+e_i$. 
Control of the relative phase of each oscillator is desired so that the wheel phases can be synchronized when climbing, as discussed in~\cite{cao2022omniwheg}, or offset when driving over flat ground for greater stability. The overall controller architecture and implementation is depicted in Fig.~\ref{f:controller}. 

The inputs to a typical differential steering model are the desired forward velocity $v$ and angular velocity $w$. An additional term $h$ is added to this controller to specify the desired height of the wheel centers from the ground. These three are taken as the command inputs to the CPG controller.
Note that each oscillator contains a frequency-related term $\omega_i$ that is controlled to affect the speed of the oscillator. The Hopf and Kuramoto models also have a parameter $\psi_{ij}$ which is controlled to affect the phase offset between oscillators.

\subsection{Kuramoto Oscillators}
A modified Kuramoto oscillator model is used as the basis for the CPG employed by \cite{crespi2008controlling}. This model allows for more direct control over the phase and amplitude of the produced oscillations. The oscillator states include the phase $\phi$, oscillation amplitude $r$, and oscillation offset $d$. The dynamics of the states are described by
\begin{align}
\dot{\phi}_i &= \omega_{i} + \sum_{i\neq j} \left(k_{ij} r_{j} \sin(\phi_j - \phi_i - \psi_{ij})\right) \nonumber \;, \\
\ddot{r}_i &= a_{r} \left( \frac{a_r}{4} (R_i - r_i) - \dot{r_i} \right) \nonumber \;,\\ 
\ddot{x}_i &= a_{x} \left( \frac{a_x}{4} (X_i - x_i) - \dot{x_i} \right) \;
\label{eq:kur}
\end{align}
where $\psi_{ij}$ is the desired phase offset between oscillators $i$ and $j$. $X_i$ and $R_i$ are the target offset and amplitude, respectively. Weights $k_{ij}$ between oscillators were all set to 1 herein. Gains $a_r$ and $a_x$ affect the convergence of $r$ and $x$ and were set as in \cite{crespi2008controlling}. We initialize the phase difference matrix $\Psi_{bias}$ so that it gives a quarter cycle offset to each oscillator, i.e.
$\Psi_{bias}(t=0) = \frac{1}{2} \pi \cdot
\left[
\begin{smallmatrix}
0 & 1 & 2 & 3\\
-1 & 0 & 1 & 2\\
-2 & -1 & 0 & 1\\
-3 & -2 & -1 & 0\\
\end{smallmatrix}
\right] $\;.
%
\par

The motor control variables, $\theta_i$ and $e_i$, are calculated to approximate the repeating U-shape trajectory of the exact inverse kinematic solution as a rectified sinusoidal signal,
\begin{equation}\label{eq:kur_out}
\begin{bmatrix}
	\theta_i\\e_i
\end{bmatrix}
=
\begin{bmatrix}
\frac{2}{N_a} \cdot \phi_i \\
r_i a_{ei} \cdot |\sin(\phi_i)| +x_i
\end{bmatrix} \;.
\end{equation}
%
To realize differential steering with the CPG the phase of each oscillator should be made to track the trajectory of a wheel in a typical differential drive system. The frequencies and phase biases are determined by 

\begin{equation}\label{eq:diff1}
\omega_{i}^* = \frac{N_a}{2} (w \frac{L_1}{h} \pm \frac{v}{h}) \;,\hspace{5pt}
\dot{\Psi}_{bias} = 2 \frac{N_a}{2} (w \frac{L_1}{h} \cdot \Psi_{ccw})\enspace,
\end{equation}
where $N_a$ is the number of arcs in the wheel and $L_1$ is the horizontal distance between wheels. With the $v$ term being negative or positive depending on which side the oscillator controls.  When turning, i.e $w\neq 0$, the desired phase difference will change over time. The phase matrix for counterclockwise turning is defined as
$\Psi_{ccw} = \left[
\begin{smallmatrix}
0 & 1 & 0 & 1\\
-1 & 0 & 1 & 0\\
0 & -1 & 0 & 1\\
-1 & 0 & -1 & 0\\
\end{smallmatrix}
\right]$.



The height input $h$ is used with inverse kinematics
~\eqref{eq:IK}
to compute the required relative phases between hubs. The oscillator offset $X_i = \phi_{O_{max}}-\phi_{I_{max}}$ is set as the relative phase when the contact point is furthest from the wheel center (i.e. maximum extension). The oscillator amplitude $R_i = X_i-(\phi_{O_{min}}-\phi_{I_{min}})$ is set as the difference between this position and the minimum extension when the contact point is directly below the wheel center. 
This creates a trajectory oscillating between the minimum and maximum extension values as the wheel turns forward.

The oscillator frequency $\omega_i$ in the Kuramoto model is applied directly to the oscillator phase resulting in sharp changes to wheel velocity when it is modified. Thus, the frequency command is filtered via dynamics
$\dot{\omega}_{i} = k_{\omega}(\omega_{i}^* - \omega_{i})$, with $k_{\omega}$  being a positive gain value. 

\subsection{Modified Hopf Oscillators}
\par
The Hopf oscillator model has been used extensively in the design of central pattern generators, particularly for legged robots. The CPG described in \cite{santos2011gait,liu2017hopf} adds an additional term to each oscillator to control the phase relationship between it and the rest of the network. This modification allows the desired phase offsets to be set in a similar way to that of the Kuramoto network discussed above, using the same phase difference matrix $\Psi_{bias}$. 

The full model for the modified Hopf oscillator is described in \eqref{eq:hop_state} below; $\omega_i$ is the frequency parameter, $\mu_i$ is the desired amplitude of the limit cycle. $a_i$ is a gain that controls the convergence rate onto the limit cycle as in \cite{liu2017hopf}. 
\begin{equation}\label{eq:hop_state}
\begin{aligned}
\begin{bmatrix}
	\dot{x}_i\\\dot{y}_i
\end{bmatrix}
=
\begin{bmatrix}
a_i (\mu_i^2 - r_i^2) & -\omega_i\\
\omega_i& a_i (\mu_i^2 - r_i^2)
\end{bmatrix}
\begin{bmatrix}
	x_i\\y_i
\end{bmatrix}
+ s_i
\\
\text{with }
s\_i = \sum_{j\neq i} k_{ij} \cdot
\begin{bmatrix}
   	\cos(\psi_{ij}) & -\sin(\psi_{ij}) \\
   	\sin(\psi_{ij}) & \cos(\psi_{ij})
\end{bmatrix}
\begin{bmatrix}
   	x_j\\y_j
\end{bmatrix}
\\
\text{and }
r_i = \sqrt{x_i^2+y_i^2}
\end{aligned}
\end{equation}

\par
The modified Hopf oscillator proposed in \cite{chen2017turboquad} correlates the state of the oscillator to the position of the wheel tip. However this came with a high computational cost to continuously solve the inverse kinematics.
Thus, a different approach was taken by using the model proposed in \cite{liu2017hopf}, which is similar to that in the Kuramoto model. The sinusoidal trajectory of the oscillator state $x_i$ is used to generate the periodic wheel extension command signal per~\eqref{eq:hop_out} below, where $a_{ei}$ is the multiplier used to scale this output to the desired hub phase difference. Thus, we have
\begin{equation}\label{eq:hop_out}
\begin{bmatrix}
	\theta_i\\e_i
\end{bmatrix}
=
\begin{bmatrix}
\frac{2}{N_a} \cdot \phi_i \\ r_i - \frac{1}{2}a_{ei}|x_i|
\end{bmatrix} \text{where }
\phi_i = \atan2(y_i,x_i)\enspace.
\end{equation}
The desired oscillator frequencies $\omega_i$ and bias dynamics $\dot{\Psi}_{bias}$ are calculated via~\eqref{eq:diff1},  
The amplitude $\mu_i$ and offset $a_{ei}$ set from the inverse kinematics as in the Kuramoto model.

\subsection{Van der Pol Oscillators}
\par
A relaxation oscillator is considered to leverage the non-sinusoidal output to more closely match the inverse kinematic trajectory. One commonly used model in CPG controllers for quadrupeds is the Van der Pol oscillator. The model proposed in~\cite{liu2009coupled}
adds a coupling term $x_{ai}$ to excite or inhibit the activity of an oscillator based on the states of the network. This has been verified to produce consistent phase relationships between oscillators. The coupling weights $k_{ij}$ are constant and defined in the matrix $K_{walk}$, that is
\begin{align}
\begin{bmatrix}
	\dot{x}_i\\\dot{y}_i
\end{bmatrix}
&=
\begin{bmatrix}
	y_i \\ a_i(p_i^2-x_i^2)\dot{x}_i - \omega_i^2 x_{ai}
\end{bmatrix}, 
\;
x_{ai} = x_i + \sum_{i \neq j} k_{ij} x_j 
\nonumber\\
K_{walk} &=
\left[
\begin{smallmatrix}
0 & -0.2 & -0.2 & -0.2\\
-0.2 & 0 & -0.2 & -0.2\\
-0.2 & -0.2 & 0 & -0.2\\
-0.2 & -0.2 & -0.2 & 0\\
\end{smallmatrix}
\right] \;. 
\label{eq:vdp}
\end{align}

A challenge to address (as in~\cite{liu2009coupled}) is the assignment of model parameters $\omega,a,p$. While the oscillator frequency is primarily determined by $\omega_i$ the effect is not linear and is affected by the other parameters. Additionally, the weight matrix can not create precise phase differences as it describes only the inhibiting or exciting influence of each oscillator on the others. Thus, continuous turning commands cannot be generated as in the other CPGs. Because of this the CPG is only compared in straight line tests with constant velocity using fixed parameters ($p_i^2=2.0, a_i=1.5, \omega_i^2 = 5.0$) determined empirically to produce a stable output close in shape to the inverse kinematics solution shown in Fig.~\ref{f:wheel}.
The maximum extension value $e_i^*$ and extension difference $a_{ei}^*$ are set directly from the inverse kinematics. The oscillator states are only used to control the phase of the wheel and the variable component of the extension commands, thus,
\begin{equation}\label{eq:vdp_out}
\begin{bmatrix}
	\theta_i\\e_i
\end{bmatrix}
=
\begin{bmatrix}
\frac{2}{N_a}\cdot \phi_i \\
e_i^* - \frac{1}{2p_i^2} (a_{ei}^* |x_i|)
\end{bmatrix}
\text{where }
\phi_i = \atan2(y_i,x_i)\;.
\end{equation}
\vspace{-15pt}



\section{Experimental Testing and Results}
\label{tests}

Three types of tests were considered. These include: 1) moving forward in a straight line while attempting to maintain a level height (physical experiments);
2) traversing uneven terrain in both straight line and curved motion (simulated testing); and 3) climbing over obstacles (both physical experiments and simulated testing).
The controller architecture and implementation is depicted in Fig.~\ref{f:controller}.
Robot pose information in physical experiments was provided by a 10-camera Optitrack motion capture system at 100Hz.
%
%
Position commands are sent to the motors from the CPGs according to~\eqref{eq:kur_out},~\eqref{eq:hop_out}, and~\eqref{eq:vdp_out} at a fixed rate of 50\,Hz.

\subsection{Experimental Flat Ground Testing}

\begin{figure}[!tb]
\vspace{6pt}
\centering
\includegraphics[width=0.499\textwidth]{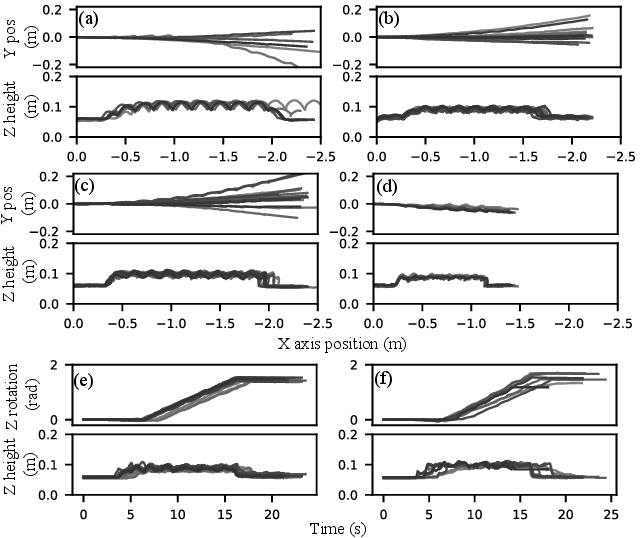}
\caption{Robot trajectory over flat ground tests. (a-d) Horizontal deviation from the commanded straight line. (e-f) Orientation change over time. The tests were performed using (a) direct drive, and (b, e) Hopf, (c, f) Kuramoto and (d) Van der Pol oscillators.}
\label{f:path_real}
\vspace{-10pt}
\end{figure}

The first group of tests consisted of the robot following a straight line trajectory on flat ground. The main goal of these tests was to determine if the periodic extension of the legs reduced the periodic vertical motion experienced when using spoked wheels. 
Additional trials were conducted with the robot turning in place. 20 trials for each of the six tests were performed. The wheels begin collapsed and are opened during the test to demonstrate the transition between rolling and walking. 
Results of these tests are shown in Fig.~\ref{f:path_real}.

According to the metrics reported in Table~\ref{t:flat_var}, the Van Der Pol model performed well in terms of reducing periodic vertical motion while also having the least deviation from a straight line trajectory. 
However, on the chosen hardware, the model became unstable when at the frequencies used in the other models. As position estimation was not considered in these controllers, all tests were run with commands sent at fixed times. Because of this, the Van Der Pol model covered less distance than other models. 
The alternating gait tended to put high loads on a single wheel and may have contributed to the drift in the $-X$ direction not seen in other models.

The Hopf and Kuramoto oscillators also succeeded in reducing the periodic vertical motion seen in the direct drive tests, but were slightly worse at following a commanded straight-line trajectory. The position of the motors as measured by the onboard encoders was also recorded for each test. In general these data closely matched commanded positions from the CPG. Observed deviations in robot height and trajectory straightness were more likely caused by wheel slipping. 
This highlights the need for higher dimensional models to close feedback control~\cite{karydis2016navigation} and planning~\cite{karydis2014planning} loops when uncertainty at ground-leg interactions can no longer be addressed by underlying models~\cite{karydis2015probabilistically} or low-level controllers. 

\begin{table}[!tb]
\vspace{6pt}
	\caption{Results from Robot Flat Ground Tests}
	\vspace{-9pt}
	\label{t:flat_var}
	\begin{center}
	\renewcommand{\arraystretch}{1.5}
	\begin{tabular}{p{1.5cm}>{\centering}p{1cm}>{\centering}p{1cm}>{\centering}p{1cm}>{\centering}p{1.8cm}}
    	\toprule
    	Oscillator Type
    	& $h$ Avg. (cm)
    	& $h$  S.D. (cm) 
    	& $v$ Avg. (cm/s)
    	& Final Pos.* \quad (100\% $\cdot$ Y/X)\\ 
    	\midrule
        Direct drive  & 11.4 & 1.20 & 16.1 & +1.85 / -9.91  \\
    	Kuramoto  	  & 10.9 & 0.60 & 13.6 & +13.3 / -4.44\\
    	Hopf     	  & 10.2 & 0.58 & 13.9 & +7.15 / -2.66  \\
    	Van der Pol   &  9.4 & 0.47 & 9.3  & -2.5 / -4.34\\
    	\bottomrule
   	 
	\end{tabular}
	\end{center}
	\footnotesize{Avg.: average; S.D.: standard deviation; Pos.: position\\
                *Final Y pos. normalized to X distance traveled}
	\vspace{-18pt}
\end{table}

\begin{figure*}[!tb]
\centering
\includegraphics[width=1\textwidth]{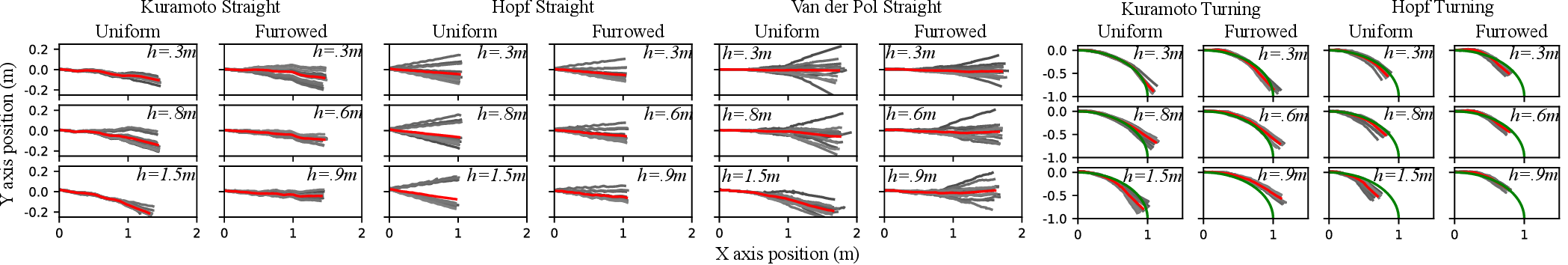}
\vspace{-15pt}
\caption{Robot paths in straight-line and turning tests on uneven terrain. Three height scales, $h$, for each combination of oscillator and terrain type were simulated. The individual test trajectories are drawn in gray, with the average position over time in red. The green path on the turning tests represents the ideal turning radius of 1m for the differential drive inputs.}
\vspace{-6pt}
\label{f:path_sim_flat}
\end{figure*}

\begin{figure*}[!tb]
\centering
\includegraphics[trim={0cm 0cm 0.0cm 0cm},clip,width=1\textwidth]{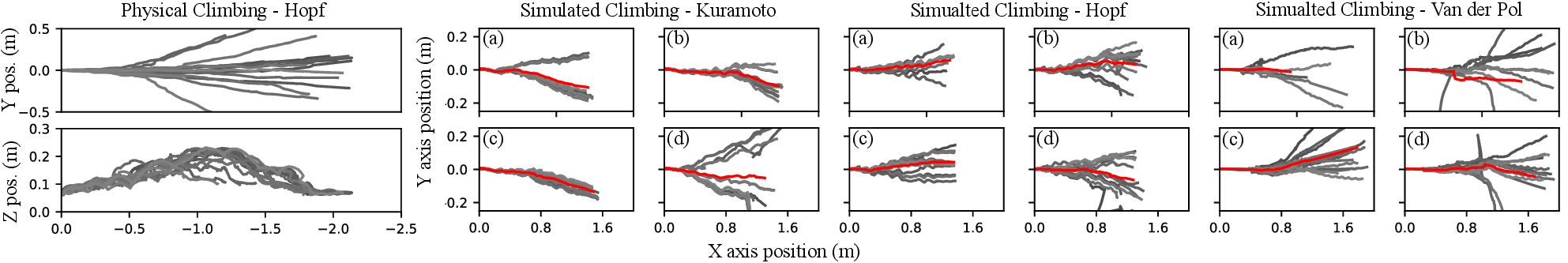}
\vspace{-15pt}
\caption{Robot paths for both physical and simulated climbing tests. The real-world experiments and first group of simulation tests (a) involved climbing a 15cm step-like block. Further simulations tests involved: (b) a cylindrical obstacle with a 10cm diameter emulating a large irrigation pipe; (c) a scattered arrangement of irregular solids emulating a field of large rocks; (d) both the pipe and rocks.}
\label{f:path_sim_climb}
\vspace{-16pt}
\end{figure*}

\subsection{Simulation Tests on Uneven Terrain}

\begin{table}[!t]
\vspace{6pt}
	\caption{Variance of Final Position From Simulation Tests (cm\textsuperscript{2})}
	\vspace{-9pt}
	\label{t:sim_var}
	\begin{center}
	\renewcommand{\arraystretch}{1.5}
	\begin{tabular}{p{1.15cm}|p{1.0cm}>{\centering}p{1.0cm}>{\centering}p{1.0cm}|p{0.9cm}>{\centering}p{0.9cm}}
	\toprule
	Terrain Type
	& Straight Kura
	& Straight Hopf
	& Straight V.D.P.
	& Turn Kura
	& Turn Hopf \\
	\midrule
   uniform 1 &122.14& 96.65& 138.75 & 22.43& 28.17\\
   uniform 2 &265.00& 158.64& 145.04& 47.00& 35.32\\
   uniform 3 &494.26& 204.58& 412.85& 113.51& 57.09\\
   wavy 1 &121.59& 74.22& 60.26& 38.41& 17.94\\
   wavy 1 &94.43& 68.07& 85.51& 54.08& 19.17\\
   wavy 1 &29.81& 64.41& 84.12& 63.30& 14.45\\
	\bottomrule
   	 
	\end{tabular}
	\end{center}
	\footnotesize{Kura: Kuramoto. V.D.P.: Van der Pol.; variance calculated relative to $y=0$ for straight line tests and to average final position in turning}
	\vspace{-21pt}
\end{table}

To enhance physical robot testing on flat ground, several tests on uneven terrain were also performed using a simulated robot model and environment. Terrain was generated using a Perlin noise function: three with uniform noise, and three scaled asymmetrically to produce furrow-like features perpendicular to the direction of travel. Straight line tests were performed with the Kuramoto, Hopf, and Van der Pol oscillator based CPGs. Because of the challenges discussed before in implementing differential steering, only the Hopf and Kuramoto based CPGs were used to test the robot's turning ability. Twelve trials for each combination of terrain and CPG were performed, with the starting phases of each wheel oscillator pair set randomly.
Figure~\ref{f:path_sim_flat} shows obtained paths and Table~\ref{t:sim_var} shows performance measured by the variance of the robot's final position.

The Kuramoto CPG appeared to have the most consistent performance across the randomized trials. The Hopf oscillator covered less distance in the same amount of time while also exhibiting a wider spread of trajectories. A lower effective frequency than the set $\omega_i$ was noticed in the Hopf oscillator when initially designing the CPG networks. This appeared to resolve when the inter-oscillator weights were turned off. The oscillators in the Van der Pol network only achieve their desired phase relationships after a rather long setting time. During this period the robot follows a straight line trajectory very well, suggesting that a synchronized gate can more effectively keep the robot moving straight.

\subsection{Obstacle Climbing Tests}
Several obstacle climbing tests were also carried out on the physical prototype and in simulation. The primary goal of these tests was to confirm the ability of the robot to traverse large obstacles. Initial tests were done with the prototype controlled by the Hopf oscillator based CPG. The robot was to climb a 15cm step-like obstacle with the wheels almost fully extended. The trajectories recorded from these tests, as well as simulation tests with other diverse obstacles are shown in Fig.~\ref{f:path_sim_climb}. The results of the step-climbing tests verified that the simulated model behaved similarly to the real robot.  Overall, the robot was able to negotiate the obstacles successfully. However, there were a few instances where the robot veered severely to one side when only one wheel made contact while climbing the step. This highlights the importance of the relative phase difference in climbing~\cite{cao2022omniwheg}. 

\vspace{-2pt}
\section{Conclusion}
\vspace{-2pt}
This paper developed a transformable leg-wheel robot design and CPG-based controllers to enable the robot to negotiate a range of terrains that can occur in semi-natural environments such as those in managed agricultural land. The paper investigated the use of central pattern generators to control mobile leg-wheel robots using generalized leg-wheel kinematics. Modifications made to CPG models provided differential steering functionality, and the oscillator states were also used to smoothen robot motion over flat terrain. 
CPG control was successful in driving the physical prototype and simulated model over a variety of terrains and obstacles.
In future work the design of the drive mechanism could be improved to provide greater rigidity and efficiency. 
The use of modified or new models in CPG control can also be explored to improve the ability of the robot to climb specific types of obstacles. 
Robots with other leg-wheel mechanisms could be used with the proposed control software
to evaluate the ability to control generalized leg-wheel mechanisms.

\bibliographystyle{IEEEtran}

\bibliography{
reference.bib
}

\end{document}